\documentclass[runningheads]{llncs}
\usepackage[T1]{fontenc}
\usepackage{graphicx}
\usepackage{amsmath}
\usepackage{amsfonts}
\usepackage{amssymb}
\usepackage[listings]{tcolorbox}

\usepackage{enumitem}
\usepackage{graphicx}

\begin{document}
%
%\title{An Empirical Study of Quantum vs. Classical Machine Learning: State of the Art}
\title{Quantum vs. Classical Machine Learning: A Unified Empirical Comparison}
%
%\titlerunning{Abbreviated paper title}
% If the paper title is too long for the running head, you can set
% an abbreviated paper title here

\author{Chuanming Yu\inst{1} \and
Jiaming Liu\inst{1} \and
Zihao Ge\inst{1} \and
Xiongfei Wu\inst{3} \and
Lulu Zhu\inst{2} \and
Pengzhan Zhao\inst{1} \and
Jianjun Zhao \inst{2}}

\authorrunning{C. Yu et al.}
% First names are abbreviated in the running head.
% If there are more than two authors, 'et al.' is used.
%
\institute{Hebei Normal University, Shijiazhuang 050024, China\\
\email{zhaopengzhan@hebtu.edu.cn}\\
\and
Kyushu University, Fukuoka 819-0395, Japan\\
\and
University of Luxembourg, Kirchberg L-1359, Luxembourg}

\maketitle              % typeset the header of the contribution
\begin{abstract}
%The abstract should briefly summarize the paper's contents in 150--250 words.
Quantum computing has emerged as a promising computational paradigm for machine learning (ML), with the potential to offer computational advantages over classical approaches. At this stage, the evidence supporting the performance and advantages of quantum machine learning (QML) models relative to classical models is insufficient. To address this gap, this paper presents an empirical study on the performance of QML models and their classical counterparts.
% We collect and compare seven sets of models, comprising supervised and reinforcement learning models.
We compare seven model pairs spanning supervised learning and reinforcement learning.
% Our research indicates that quantum machine learning models have not yet surpassed classical models in generalization, particularly in accuracy, policy stability, and training efficiency.
Our results indicate that the evaluated quantum machine learning models do not yet surpass the classical baselines in overall prediction performance, policy stability, or training time.
Nevertheless, QML remains a promising approach for filtering noise and controlling false positives. 
Our research findings summarize the challenges facing quantum machine learning across hardware environments, training efficiency, and convergence stability, providing a foundation for research into the robustness and parameter optimization of QML.
This work is publicly available at https://github.com/Z-537-437/QML.

\keywords{Quantum Machine Learning \and Generalization Ability \and Empirical Study.}
\end{abstract}

\section{Introduction}
Quantum computing has made substantial progress due to advances in computing power and algorithms~\cite{cerezo2021}. Tremendous efforts from industry and academia have greatly stimulated the evolution of this area. Although fault-tolerant quantum computers are likely not available in the near future, research on quantum machine learning has revealed the potential of quantum computers to outperform classical computers in machine learning tasks~\cite{Biamonte_2017}. 
With advances in quantum machine learning, numerous QML algorithms have been proposed~\cite{havlicek2019supervised,zoufal2019quantum}. Currently, specialized quantum machine learning frameworks have further driven the convergence of quantum computing and machine learning, making it easier to design and train learning models in a quantum environment~\cite{bergholm2018pennylane,meyer2024qiskit}. Against this backdrop, various quantum neural networks (QNNs) have been developed, such as QCNN~\cite{cong2019quantum} and QLSTM~\cite{chen2025toward}.
Given the rapid evolution of this field and the practical applications of quantum machine learning, it is particularly important to verify the generalization capabilities of QML models. Numerous studies examine the performance of QML models in practical applications~\cite{sharna2024quantum,tasnim2025quantum}, such as financial forecasting~\cite{ahmad2026quantum}, medical image classification~\cite{tasnim2024comparison}, and computational chemistry~\cite{chakraborty2025classical}. However, it remains unclear whether QML models offer any advantages in overall generalization compared to traditional models.

To bridge this gap, we propose an empirical study that combines supervised and reinforcement learning to quantify the differences between QML models and traditional ML models. Specifically, we evaluate seven representative model pairs across classification tasks %(using the Bars and Stripes dataset) 
and sequential decision-making tasks. The key findings of this study include the following:
%(within a custom Hypercube Environment).
%A core aspect of our methodology is the strict adherence to an "Architecture and Inductive Bias Matching" principle, ensuring that the classical baselines are restricted to the same parameter magnitudes as their quantum counterparts to provide a rigorous and fair comparison. 
\begin{itemize}
    \item Classical machine learning (CML) models currently outperform the evaluated QML models in terms of overall accuracy, policy stability, and training efficiency across both classification and sequential decision-making tasks.
    
    \item Despite trailing in overall recall, certain QML architectures demonstrate distinct advantages in filtering noise and false-positive control, such as the Quantum Convolutional Neural Network (QCNN), achieving notably higher precision. %Furthermore, parameterized quantum circuits show parameter-efficient decision-making capabilities competitive with exact classical tabular methods

    \item We identify four critical challenges hindering the advancement of QML: qubit bottlenecks that necessitate strict dimensionality reduction, vulnerability to hardware noise, significant gaps in training efficiency, and optimization instability, including barren plateaus.
\end{itemize}

In summary, this paper contributes the following:
\begin{itemize}
    \item The first systematic empirical comparison between QML and CML models covering seven model pairs across both supervised and reinforcement learning tasks under a unified framework.

    \item A rigorous evaluation methodology based on architectural alignment that isolates the true representational power of quantum models from the parameter-scale advantages of classical models.

    \item Insights into the specific operating conditions where QML currently exhibits advantages and a detailed summary of the challenges researchers face across hardware environments, noise interference, and convergence stability.
\end{itemize}

The rest of the paper is organized as follows. Section~\ref{sec:related-work} reviews related work in quantum machine learning and existing benchmarking studies. Section~\ref{sec:methodology} details our methodology, including datasets, environments, architectures, and evaluation metrics. Section~\ref{sec:empirical-results} presents the empirical results and answers our research questions. Section~\ref{sec:conclusion} finally concludes this paper.

\section{Related Work}
\label{sec:related-work}

Previous research has provided a comprehensive review of quantum machine learning~\cite{akrom2024quantum,adebayo2023variational}. Rodríguez-Díaz et al. provide a comprehensive survey of the foundations, algorithms, frameworks, data, and applications of QML, serving as an important reference for researchers and practitioners~\cite{rodriguez2025survey}.
In a similar vein, Lamichhane and Rawat review recent advances, challenges, and future perspectives of QML, with particular attention to its applications in related domains~\cite{lamichhane2025quantum}.
Devadas and Sowmya summarized the history of QML, whilst highlighting the issues that remain unresolved at this stage ~\cite{devadas2025quantum}.
% This study is an empirical investigation that examines the performance of QML and ML in the absence of noise and in the presence of weak noise.
Unlike the above surveys, this work empirically compares QML and classical ML under noiseless and limited-noise settings, where applicable.

Existing benchmark and empirical studies examine the performance of QML in multiple fields.
Ahmad et al. proposed a reproducible benchmarking framework for financial forecasting and noted that QML does not outperform classical models in all scenarios~\cite{ahmad2026quantum}.
Kruse et al. proposed evaluation metrics for evaluating quantum reinforcement learning models~\cite{kruse2025benchmarking}.
Drawing on PennyLane and Fujitsu Qulacs, Adnan compared the performance of quantum methods, which focus on metrics such as accuracy, computation time, resource consumption, and scalability~\cite {adnan2025competitive}.
Our work covers typical QML algorithms for supervised and reinforcement learning, intending to provide a comprehensive examination of how QML currently performs compared to traditional ML.

\section{Methodology}
\label{sec:methodology}

\begin{figure}[htbp]
    \centering
    \includegraphics[width=0.96\textwidth]{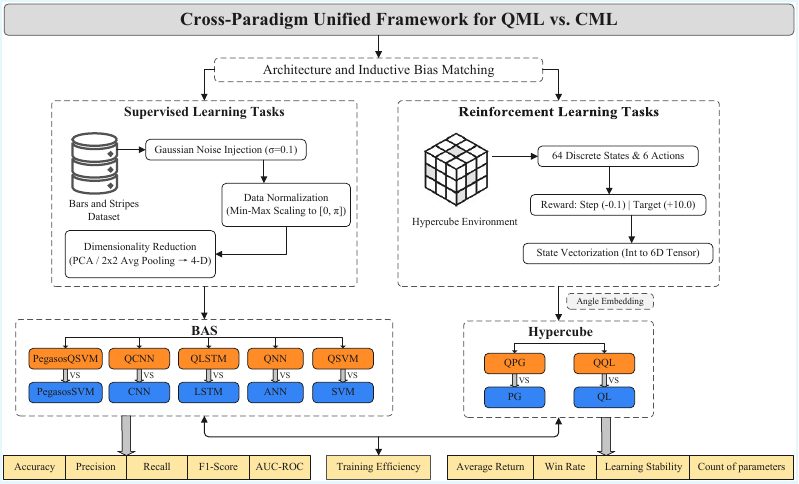}
    \caption{Cross-Paradigm Unified Framework for QML vs. CML}
    \label{fig:flowchart}
\end{figure}

In this section, we describe how we conducted our study, including formulating research questions, selecting QML models, constructing the dataset, and conducting the study. The architecture of our approach is shown in Figure~\ref{fig:flowchart}.
\subsection{Research Question}
The following three research questions drive this study:
\begin{itemize}[leftmargin=2em]
\setlength{\itemsep}{3pt}
  % \item We distill a set of common bug patterns for quantum programs, which is the foundation of performing static analysis on quantum programs.
  \item \textit{RQ1: How do quantum machine learning models compare with traditional models in terms of performance?}
  This question serves as our first purpose: to assess whether the generalization capabilities of quantum machine learning models have surpassed those of traditional models at this stage.
  
  \item \textit{RQ2: What are the advantages of quantum machine learning models, and what conditions are required?}
  This question is the core and foundation of this study. Understanding the current strengths and limitations of quantum machine learning will clarify its practical applicability and provide a foundation for designing more robust and effective approaches.

  %\item \textit{RQ3:What challenges do users face when using QML frameworks?}
  %Due to the multi-platform and multi-device implementation of QML, users are faced with more than simply using the QML platforms.
  %\item \textit{RQ3:What are the common challenges for developers when developing QML frameworks?}
  \item \textit{RQ3:What challenges remain in advancing the development of quantum machine learning models?}
    The generalization ability of quantum machine learning models is affected not only by noise in quantum hardware but also by robustness issues common to classical models.
\end{itemize}
\subsection{Experimental Objectives and Cross-Paradigm Framework}
Our study aims to systematically compare the performance and parameter efficiency of Quantum Machine Learning (QML) and Classical Machine Learning (CML) models across various learning tasks under a unified experimental framework. Specifically, this work covers both supervised learning and reinforcement learning to provide a comparative analysis of the performance landscapes and operational boundaries of both quantum and classical paradigms.

In classification tasks, we examine the impact of quantum encoding on model performance. To this end, a Bars and Stripes (BAS) dataset, derived from the canonical benchmark in the QML community, is selected to assess whether QML models can more effectively represent and classify data when patterns are embedded in a quantum state space.

In reinforcement learning tasks, we construct a unified discrete environment to contrast the learning trajectories and convergence properties of Quantum Reinforcement Learning (QRL) and Classical Reinforcement Learning (CRL) models in sequential decision-making problems.

To ensure fairness and reliability, we establish a unified benchmarking framework implemented in PennyLane, following these principles:

\begin{itemize}
    \item \textbf{Task Consistency:} All classification models are evaluated on the same BAS dataset, while reinforcement learning models interact within the same Hypercube Environment.
    \item \textbf{Consistent Protocol:} All tasks within the same domain share identical execution rules in order to ensure rigorous intra-category fairness. Specifically, all classification tasks adopt a unified training/test split, while reinforcement learning tasks share identical environmental constraints and reward structures, with performance evaluated via a consistent rolling window.
    \item \textbf{Unified Evaluation:} In classification tasks, there is a comprehensive set of metrics, including Accuracy, Precision, Recall, F1-Score, and AUC-ROC. In reinforcement learning, cumulative rewards and stability metrics computed during the convergence phase are used.
\end{itemize}

Based on this framework, we systematically analyze quantum and classical models in two dimensions with respect to their well-suited data distributions and their sequential decision-making capabilities. This approach provides a nuanced understanding of how differences in model paradigms affect learning performance and resource utilization.

\subsection{Datasets and Environments}

\subsubsection{\textit{Bars and Stripes (BAS) Dataset}.}

The Bars and Stripes (BAS) dataset is used as a benchmark for classification because its structured patterns are well-suited for quantum encoding. Each sample consists of an $N \times N$ binary image, flattened into a feature vector representing either horizontal bars or vertical stripes. To simulate real-world imperfections and prevent overfitting, adjustable Gaussian noise (parameterized by the standard deviation $\sigma$) is added to the images before normalization. Furthermore, for specific quantum models constrained by the number of available qubits (e.g., QSVM), dimensionality reduction techniques such as Principal Component Analysis (PCA) can be applied to project features into a lower-dimensional space. This controlled synthetic environment allows us to isolate and compare the classification capabilities of the models across comprehensive evaluation metrics, including F1-Score, AUC-ROC, and computational efficiency.

\subsubsection{\textit{Hypercube Environment}.}

To evaluate the proposed methods in a sequential decision-making setting, we introduce the Hypercube Environment, a custom Gymnasium-compliant environment specifically designed and implemented for this purpose. The state space is defined by the vertices of an $n$-dimensional hypercube ($n = 6$ in our experiments), where each vertex corresponds to a unique classical binary string of length $n$. This structure enables a direct mapping onto the quantum state space via angle embedding, in which each bit of the binary string is encoded as the rotation angle of the corresponding qubit. Specifically, the environment is initialized to the classical all-zero string, and the agent aims to reach the all-one target string, represented by a fully rotated state configuration. Each non-terminal step receives a penalty of $-0.1$, while reaching the target state grants a terminal reward of $+10.0$, after which the episode terminates. Episodes are also truncated after at most $T = 40$ steps to ensure computational efficiency. This standardized environment serves as a rigorous testbed for evaluating the average return, final win rate, and learning stability of various quantum reinforcement learning algorithms relative to their classical counterparts.

\subsection{Algorithms and Architecture Alignment}
\label{Algorithms}
In this paper, we use the PennyLane framework and the PyTorch deep learning interface, seamlessly integrating with the classical baselines implemented in PyTorch and Scikit-learn. Given the limitations of current fault-tolerant quantum hardware, all quantum circuits are evaluated on a noiseless state-vector quantum simulator, and kernel methods are used to accelerate the computation of the quantum-state fidelity kernel matrix using multi-core parallel computing. All experiments are executed on classical computers and the global random seed is fixed (\texttt{seed = 42}) to improve reproducibility.
%To eliminate potential interference from model capacity discrepancies in performance evaluation, this study strictly adheres to the principle of "Architecture and Inductive Bias Matching". 
On the other hand, the hyperparameter settings and network architectures for all experiments are strictly aligned to maintain comparable parameter scales. Specifically, we deliberately restrict the internal complexity of the classical baselines to be on the same order of magnitude as that of their quantum counterparts. This ensures a fair performance comparison rather than relying on the massive parameter-scale advantages of classical models.

\subsubsection{Supervised Learning Configurations}

\paragraph{\textbf{Quantum Supervised Learning Models.}} In the Bars and Stripes (BAS) image classification task, the quantum models are designed to operate within the strict constraints of a 4-qubit simulation environment. To accommodate the periodicity of quantum angle embedding, 16-dimensional input features from the 4×4 pixel grids are normalized to the $[0, \pi]$ range using MinMaxScaler. Dimensionality reduction is then applied to address the qubit bottleneck: the QNN and QSVM models use Principal Component Analysis (PCA) to reduce the feature dimensionality to 4 dimensions. At the same time, the QCNN employs $2 \times 2$ average pooling to downsample the data while preserving spatial relationships. Architecturally, the QNN consists of an angle embedding and six BasicEntanglerLayers, and the QCNN features five custom convolutional blocks (RY rotations and CNOT gates) totaling approximately 30 parameters. For sequential data, the QLSTM employs two entangler layers for hidden-state transitions and incorporates a Depolarizing Channel to simulate hardware noise. Furthermore, QSVM variants leverage ZZFeatureMaps with linear or all-to-all entanglement to exploit the high-dimensional expressive power of the Hilbert space.

\paragraph{\textbf{Classical Supervised Learning Models.}}
%To ensure a rigorous and fair comparison, the classical models were deliberately "downsized" to align their capacity with the limited parameter count of the quantum architectures. 
We specifically reduce the dimensionality of the classic model to ensure the rigor and fairness of the comparison.
The ANN is simplified to a minimal two-hidden-layer network (16 and 8 neurons), whereas the CNN is restricted to a single 6-channel convolutional layer followed by a linear output. For recurrent tasks, the LSTM hidden state dimension is capped at $h=8$, and an early-stopping mechanism (patience=10) is implemented to prevent overfitting on the synthetic dataset. Training is conducted using the Adam optimizer and Mean Squared Error (MSE) loss, with additive Gaussian noise ($\sigma = 0.1$) injected to assess robustness in real-world simulations. In the realm of kernel methods, the Pegasos-SVM and standard SVM use RBF kernels and are optimized via stochastic gradient descent with L2 regularization. This low-capacity configuration serves as a critical baseline, revealing the performance boundaries of classical algorithms when restricted to the same scale as current NISQ-era quantum models.

\vspace{1em}

\subsubsection{Reinforcement Learning Configurations}
\paragraph{\textbf{Quantum Reinforcement Learning Models.}} The quantum agents are evaluated within a 6-qubit Hypercube Environment featuring 64 discrete states and 6 possible actions. For policy-based learning, the Quantum Policy Gradient (QPG) agent utilizes a parameterized quantum circuit (PQC) in which state information is encoded via rotation-based state preparation, followed by 6 layers of strongly entangling gates comprising universal single-qubit rotations and entangling operations. The circuit outputs Pauli-Z expectation values, which are scaled by a learnable temperature parameter (initially set to 5.0) and processed through a Softmax function to determine action probabilities. For value-based learning, the Quantum Q-Learning (QQL) agent adopts a hybrid architecture comprising 4 layers of parameterized quantum entangling operations and a classical linear output layer that maps quantum expectations to action values. The QQL agent employs a $\epsilon$-greedy strategy, with the exploration rate exponentially decaying from 0.5 to 0.05 at a rate of 0.99 per episode. Both quantum models are optimized using the Adam optimizer ($lr=0.01$) over an 800-episode training horizon, with the QQL agent utilizing a 5,000-capacity replay buffer and a batch size of 16.

\paragraph{\textbf{Classical Reinforcement Learning Models.}} The classical counterparts are implemented as neural network-based function approximators to provide a benchmark for the same environment. The classical Policy Gradient (PG) agent is structured as a Multi-Layer Perceptron (MLP) with a single hidden layer of $32$ neurons and ReLU activation. It employs a discount factor $\gamma=0.95$ and performs updates via baseline-normalized advantage estimation to match the QPG’s optimization strategy. Similarly, the Classical Q-Learning (QL) baseline is developed as a Deep Q-Network (DQN) with an identical 32-neuron hidden-layer architecture. To ensure a consistent comparison with the QQL agent, this baseline utilizes the same 5,000-capacity experience replay mechanism, a mini-batch size of 16, and an identical exponential $\epsilon$-greedy decay schedule with a discount factor $\gamma=0.99$. All classical models are trained for 800 episodes using the Adam optimizer with a learning rate of 0.01 to ensure a consistent optimization landscape.

\subsection{Evaluation Metrics}
\label{subsec:metrics}
%To facilitate a rigorous and unbiased quantitative assessment of both quantum and classical paradigms, 
This work employs standardized metric arrays tailored to the specific requirements of both supervised and reinforcement learning tasks. These metrics assess not only the models’ generalization capabilities but also their training and computational efficiency.
%These metrics evaluate not only the models' predictive performance but also their computational efficiency and robustness.

\subsubsection{\textit{Supervised Learning Metrics}.}
For all classification tasks, the following metrics are recorded:
%—covering kernel-based methods (Pegasos QSVM vs. SVM, Kernel QSVM vs. SVM), feed-forward networks (QNN vs. ANN), convolutional architectures (QCNN vs. CNN), and sequential models (QLSTM vs. LSTM)—the following metrics are recorded:

\begin{itemize}
    \item \textbf{Accuracy:} The primary measure of the proportion of total samples correctly classified.
    \item \textbf{Precision and Recall:} Utilized to evaluate the model's specificity and sensitivity, particularly in identifying structured patterns within the Bars and Stripes (BAS) dataset.
    \item \textbf{F1-Score:} The harmonic mean of precision and recall, serving as a robust indicator of performance under potential class imbalances.
    \item \textbf{AUC-ROC:} Evaluates the model's discriminative capability across various decision thresholds, providing insight into the quality of the learned feature space.
    \item \textbf{Training Efficiency:} Measures the actual computational overhead and total convergence time required during the training phase, providing a practical comparison between quantum simulations and classical computations.
\end{itemize}

\subsubsection{\textit{Reinforcement Learning Metrics}.}
For decision-making tasks, a comparison is made across the following five key dimensions:
%such as the qubit-state reaching environment, the performance of \textit{Quantum Q-Learning} (QQL) and \textit{Quantum Policy Gradient} (QPG) is benchmarked against classical baselines using five key dimensions:

\begin{itemize}
    \item \textbf{Average Return:} This metric represents the mean cumulative reward obtained by the agent over the final training episodes.% defining the upper bound of the learned policy's utility.
    \item \textbf{Win Rate:} This measures the frequency with which the agent successfully reaches the target quantum state within the maximum step limit ($T=40$).
    \item \textbf{Learning Stability:} Measured by the standard deviation of episode rewards during convergence, this metric assesses the algorithm's resilience to exploration variance and optimization challenges such as Barren Plateaus.
    \item \textbf{Count of parameters:} This denotes the total number of trainable parameters in the model. By observing this metric alongside the average return and win rate, we compare the parameter efficiency across different architectures.
    \item \textbf{Training Efficiency:}This quantifies the total computational time required to complete the training process, reflecting the practical computational overhead of the algorithms.
\end{itemize}

\section{Empirical Results}
\label{sec:empirical-results}

In this section, we present the empirical results of our comparative study between Quantum Machine Learning (QML) and Classical Machine Learning (CML) models. Following the unified experimental framework described in Section~\ref{sec:methodology}, our evaluation is divided into two main categories: supervised classification tasks and reinforcement learning tasks.

The models are evaluated using the metrics defined in Section~\ref{sec:methodology}, which assess both predictive performance and learning dynamics. Based on these results, we examine the performance differences between quantum and classical models with respect to generalization, convergence, and parameter efficiency.

\subsection{RQ1:Performance of Quantum vs. classical machine learning}

We conduct a systematic evaluation of five pairs of quantum versus classical models on the BAS dataset, and of two pairs of reinforcement learning algorithms in the Hypercube Environment. All experiments strictly adhere to the unified framework and architectural alignment principles described in Section~\ref{sec:methodology}.
The detailed experimental data are presented in Table~\ref{tab:classification_results} and Table~\ref{tab:rl_results}:

\begin{table}
\caption{Classification Performance on the BAS Dataset.}\label{tab:classification_results}
\centering
% 将表格宽度限制为当前文本行宽
\resizebox{\linewidth}{!}{%
\begin{tabular}{llcccccc}
\hline
\textbf{Model Pair} & \textbf{Model} & \textbf{Accuracy} & \textbf{Precision} & \textbf{Recall} & \textbf{F1-Score} & \textbf{AUC-ROC} & \textbf{Training Eff.} \\ % 稍微缩写了 Efficiency
\hline
QSVM vs SVM & QSVM & 0.8250 & 0.8060 & 0.7826 & 0.7941 & 0.9059 & 11.4005 \\
 & SVM &  \textbf{0.9250} & \textbf{0.8904} &  \textbf{0.9420} &  \textbf{0.9155} &  \textbf{0.9861} & \textbf{0.0066} \\
\hline
QNN vs ANN & QNN & 0.9500 & 0.9506 & 0.9506 & 0.9506 & 0.9902 & 14.2969 \\
 & ANN &  \textbf{0.9812} &  \textbf{0.9643} &  \textbf{1.0000} &  \textbf{0.9818} &  \textbf{1.0000} & \textbf{0.6622} \\
\hline
PegasosQSVM vs & PegasosQSVM & 0.9062 &  \textbf{0.9610} & 0.8605 & 0.9080 & 0.9486 & 3.5874 \\
PegasosSVM & PegasosSVM &  \textbf{0.9563} & 0.9247 &  \textbf{1.000} &  \textbf{0.9609} &  \textbf{0.9928} & \textbf{0.3362} \\
\hline
QCNN vs CNN & QCNN & 0.8187 & \textbf{0.9014} & 0.7442 & 0.8153 & 0.9150 & 42.5999 \\
 & CNN & \textbf{0.8250} & 0.8537 & \textbf{0.8140} & \textbf{0.8333} & \textbf{0.9305} & \textbf{2.1457} \\
\hline
QLSTM vs LSTM & QLSTM & 0.7750 & 0.7529 & 0.8101 & 0.7805 & 0.8903 & 564.4091 \\
 & Classical LSTM & \textbf{0.8500} & \textbf{0.8235} & \textbf{0.8861} & \textbf{0.8537} & \textbf{0.9441} & \textbf{18.4982} \\
\hline
\end{tabular}%

}
\end{table}

\begin{table}
\caption{Reinforcement Learning Performance in the Hypercube}\label{tab:rl_results}
\centering
\begin{tabular}{llccccc}
\hline
\textbf{Model Pair} & \textbf{Model} & \textbf{Average Return} & \textbf{Win Rate} & \textbf{Stability} & \textbf{Params} & \textbf{Training Eff.} \\
\hline
QPG vs PG & QPG & 6.9360 & 0.8800 & 4.1365 & \textbf{109} & 2150.6730 \\
 & PG & \textbf{9.4440} & \textbf{1.0000} & \textbf{0.1061} & 422 & \textbf{6.2230} \\
\hline
QQL vs QL & QQL & 8.0780 & 0.9400 & 3.1567 & \textbf{66} & 1868.4183 \\
& QL & \textbf{9.4640} & \textbf{1.0000} & \textbf{0.0866} & 422 & \textbf{10.3986} \\
\hline
\end{tabular}
\end{table}

\subsubsection{\textit{Objective Data Analysis}.}
Based on the empirical results presented in Table~\ref{tab:classification_results} and Table~\ref{tab:rl_results}, we evaluate the macroscopic performance differences between the two paradigms:

\begin{itemize}
    \item \textbf{Comprehensive Performance Lead of Classical Models:} In both supervised classification and reinforcement learning tasks, classical machine learning (CML) models demonstrate a clear and consistent superiority. Across all classification pairings, classical baselines achieve higher overall accuracy, F1 Scores, and AUC-ROC than their quantum counterparts. Similarly, in reinforcement learning, classical agents consistently reach perfect win rates and optimal average returns.

    \item \textbf{Robustness and False Negative Rates:} A macroscopic view of the recall metrics reveals that classical models possess a stronger capability to identify positive instances correctly. The consistent gap in recall indicates that current QML architectures are generally more susceptible to false negatives in structured datasets.

    \item \textbf{Training Efficiency and Computational Overhead:} There is a fundamental disparity in computational costs. Across the board, classical models train significantly faster. The training efficiency gap spans from an order of magnitude in simpler networks to dozens of times in recurrent architectures, highlighting the severe computational overhead currently faced by QML.

    \item \textbf{Optimization and Policy Stability:} In sequential decision-making environments, classical algorithms exhibit a markedly higher stability during training. Quantum reinforcement learning agents experience significantly higher variance and uncertainty during policy convergence, making classical models far more reliable for stable training.
\end{itemize}

%\subsubsection{\textit{Summary and Answer to RQ1}.}
To directly address RQ1, our results indicate that the overall generalization capabilities of quantum machine learning models do not yet surpass those of classical models. Classical paradigms maintain a substantial lead in overall predictive accuracy, policy stability, and training efficiency across both supervised and reinforcement learning domains.

\subsection{RQ2:Advantages and Necessary Conditions of quantum machine learning}

While classical models generally maintain a macroscopic performance lead, a granular analysis of our empirical results reveals that Quantum Machine Learning (QML) possesses unique representational advantages under specific configurations. To directly address RQ2, we identify two primary advantages of QML and outline the strict operating conditions required to realize them.

\subsubsection{Specific Advantages of QML}
\begin{itemize}
    \item \textbf{Enhanced Precision and False Positive Control:} While QML generally trails in recall, specific architectural designs demonstrate superior specificity. As shown in Table 1, the Quantum Convolutional Neural Network (QCNN) achieves a notably higher precision (0.9014) compared to its classical CNN counterpart (0.8537). This indicates that quantum encoding mechanisms have distinct potential to effectively filter noise and strictly control false positives in certain spatial data distributions.
    
    \item \textbf{Parameter-Efficient Sequential Decision-Making:} Results in Table 2 highlight the superior parameter efficiency of quantum agents in the Hypercube navigation task. The Quantum Q-Learning (QQL) agent achieves a 94.0\% win rate using only 66 parameters—an 84.4\% reduction compared to the classical baseline (422 parameters, 100\% win rate). Similarly, the Quantum Policy Gradient (QPG) maintains competitive performance with only 109 parameters, versus 422 for its classical counterpart. These findings empirically demonstrate that parameterized quantum circuits possess a high representational density, enabling effective policy learning within a remarkably compressed parameter space.
\end{itemize}

\subsubsection{Necessary Operating Conditions}
\begin{itemize}
    \item \textbf{Strict Dimensionality Reduction:} Due to NISQ-era qubit limitations, features must be heavily compressed (e.g., via PCA) and normalized via angle encoding to fit the quantum Hilbert space.
    
    \item \textbf{Noise Robustness and Environment:} While shallow quantum classifiers demonstrate some noise resilience, the deep quantum reinforcement learning models in this study are evaluated on noise-free simulators to isolate and verify their theoretical representational power explicitly. Transitioning these complex networks to near-term physical hardware requires efficient noise mitigation techniques or fault-tolerant devices to maintain performance and prevent such phenomena.
    %such as "barren plateaus."
\end{itemize}

\subsection{RQ3: challenges of advancing quantum machine learning}
Building on the preceding comparative analysis, while Quantum Machine Learning (QML) has demonstrated baseline feasibility and certain localized advantages, the field continues to face formidable challenges. Synthesizing the empirical findings of this study, these obstacles are primarily categorized into four dimensions: hardware constraints, noise interference, training efficiency, and optimization stability.

\subsubsection{\textit{Qubit Bottleneck and Dimensionality Limits}.}
The nascent state of fault-tolerant quantum hardware currently tethers the development of quantum models. To process complex high-dimensional datasets, such as images, researchers are compelled to employ dimensionality reduction techniques, such as Principal Component Analysis (PCA) or average pooling layers, to mitigate the qubit bottleneck. This requisite feature compression restricts the model’s ability to directly leverage high-dimensional raw data, potentially leading to critical information loss.

\subsubsection{\textit{Hardware Noise and Robustness Issues}.}
The generalization performance of QML models is subject not only to classical robustness challenges but also intrinsically constrained by the stochastic noise inherent in quantum systems. Empirical tests with simulated hardware noise (e.g., via Depolarizing Channels in QLSTM) and Gaussian-noise injections reveal that QML models are highly sensitive to environmental imperfections. This sensitivity often precipitates performance degradation in real-world or high-noise scenarios.

\subsubsection{\textit{Significant Gap in Training Efficiency}.}
A profound disparity exists between classical and quantum training efficiencies. While quantum supervised learning models inherently require an order of magnitude more training time than classical ones (e.g., the overhead of QLSTM is over 30 times greater), this gap is drastically amplified in reinforcement learning, as evidenced by Table~\ref{tab:rl_results}. Because every environmental interaction in QRL requires executing the parameterized quantum circuit, the massive frequency of state preparations and measurements creates severe computational bottlenecks, which fundamentally accounts for the extensive training times. On real hardware, executing the multi-layered entangling circuits required for such sequential decision-making is strictly constrained by quantum decoherence times, making QRL scalability far more challenging than supervised learning.

\subsubsection{\textit{Optimization Stability and Barren Plateaus}.}
In reinforcement learning tasks involving sequential decision-making, quantum agents exhibit pronounced fluctuations and uncertainty during the policy convergence phase. The stability metrics for classical Policy Gradient (PG) significantly outperform those of Quantum Policy Gradient (QPG). These high-variance trajectories suggest that QML optimization is particularly vulnerable to the "Barren Plateaus" phenomenon and stochastic noise, thereby impeding stable convergence to optimal policies.

In conclusion, the current trajectory of QML is hindered not only by the physical constraints of the NISQ era but also by the algorithmic limitations regarding training efficiency and convergence stability. Future advancements must focus on designing noise-resilient quantum architectures and developing more robust parameter-optimization protocols.

\section{Conclusion}
\label{sec:conclusion}

In this paper, we present a unified empirical comparison of seven QML/CML model pairs across supervised learning and reinforcement learning. %We collected seven representative QML models and compared their performance with that of their corresponding ML models.
Our research indicates that while the evaluated QML models do not yet outperform the classical baselines, they show strong potential for generalization and still face challenges. 
Future work should focus on robustness, noise mitigation, scalable training, and more stable optimization strategies for QML.

%---- Bibliography ----
% BibTeX users should specify bibliography style 'splncs04'.
% References will then be sorted and formatted in the correct style.
%%\bibliographystyle{IEEEtranS}
%\bibliography{IEEEabrv, ref}
\bibliographystyle{splncs04}
\bibliography{ref}

\end{document}